\documentclass[sigconf]{acmart}

\usepackage{booktabs} 

\setcopyright{rightsretained}

\copyrightyear{2017} 
\acmYear{2017} 
\setcopyright{acmcopyright}
\acmConference{MUSA2'17}{October 27, 2017}{Mountain View, CA, USA}\acmPrice{15.00}\acmDOI{10.1145/3132515.3132520}
\acmISBN{978-1-4503-5509-4/17/10}

\begin{document}
\title{More cat than cute? \\Interpretable Prediction of Adjective-Noun Pairs}

\author{D{\`e}lia Fern{\'a}ndez}
\authornote{Work partially developed while D{\`e}lia Fern{\'a}ndez was a visiting scholar at Columbia University.}
\affiliation{%
  \institution{Vilynx}
  \city{Barcelona} 
  \country{Catalonia/Spain} 
}
\email{delia@vilynx.com}

\author{Alejandro Woodward}
\affiliation{%
  \institution{Universitat Polit{\`e}cnica de Catalunya}
  \city{Barcelona} 
  \country{Catalonia/Spain} 
}
\email{alejandro.woodward@alu-etsetb.upc.edu}

\author{V{\'i}ctor Campos}
\affiliation{%
  \institution{Barcelona Supercomputing Center}
  \city{Barcelona} 
  \state{Catalonia/Spain} 
}
\email{victor.campos@bsc.es}

\author{Xavier Gir{\'o}-i-Nieto}
\affiliation{%
  \institution{Universitat Polit{\`e}cnica de Catalunya}
  \city{Barcelona} 
  \country{Catalonia/Spain} 
}
\email{xavier.giro@upc.edu}

\author{Brendan Jou}
\affiliation{%
  \institution{Columbia University}
  \city{New York City}
  \state{New York} 
}
\email{bjou@ee.columbia.edu}

\author{Shih-Fu Chang}
\affiliation{%
  \institution{Columbia University}
  \city{New York City}
  \state{New York} 
}
\email{sfchang@ee.columbia.edu}

\renewcommand{\shortauthors}{D. Fern{\'a}ndez et al.}

\begin{abstract}
The increasing availability of affect-rich multimedia resources has bolstered interest in understanding sentiment and emotions in and from visual content. Adjective-noun pairs (ANP) are a popular mid-level semantic construct for capturing affect via visually detectable concepts such as ``cute dog" or ``beautiful landscape". Current state-of-the-art methods approach ANP prediction by considering each of these compound concepts as individual tokens, ignoring the underlying relationships in ANPs. This work aims at disentangling the contributions of the `adjectives' and `nouns' in the visual prediction of ANPs. Two specialised classifiers, one trained for detecting adjectives and another for nouns, are fused to predict 553 different ANPs. The resulting ANP prediction model is more interpretable as it allows us to study contributions of the adjective and noun components.
\end{abstract}

%
%

\ccsdesc[500]{Information systems~Multimedia information systems}
\ccsdesc[500]{Computing methodologies~Scene understanding}

\keywords{affective computing, convolutional neural networks, compound concepts, adjective noun pairs, interpretable models}


\maketitle

\section{Introduction}

\begin{figure}
\includegraphics[width=\linewidth]{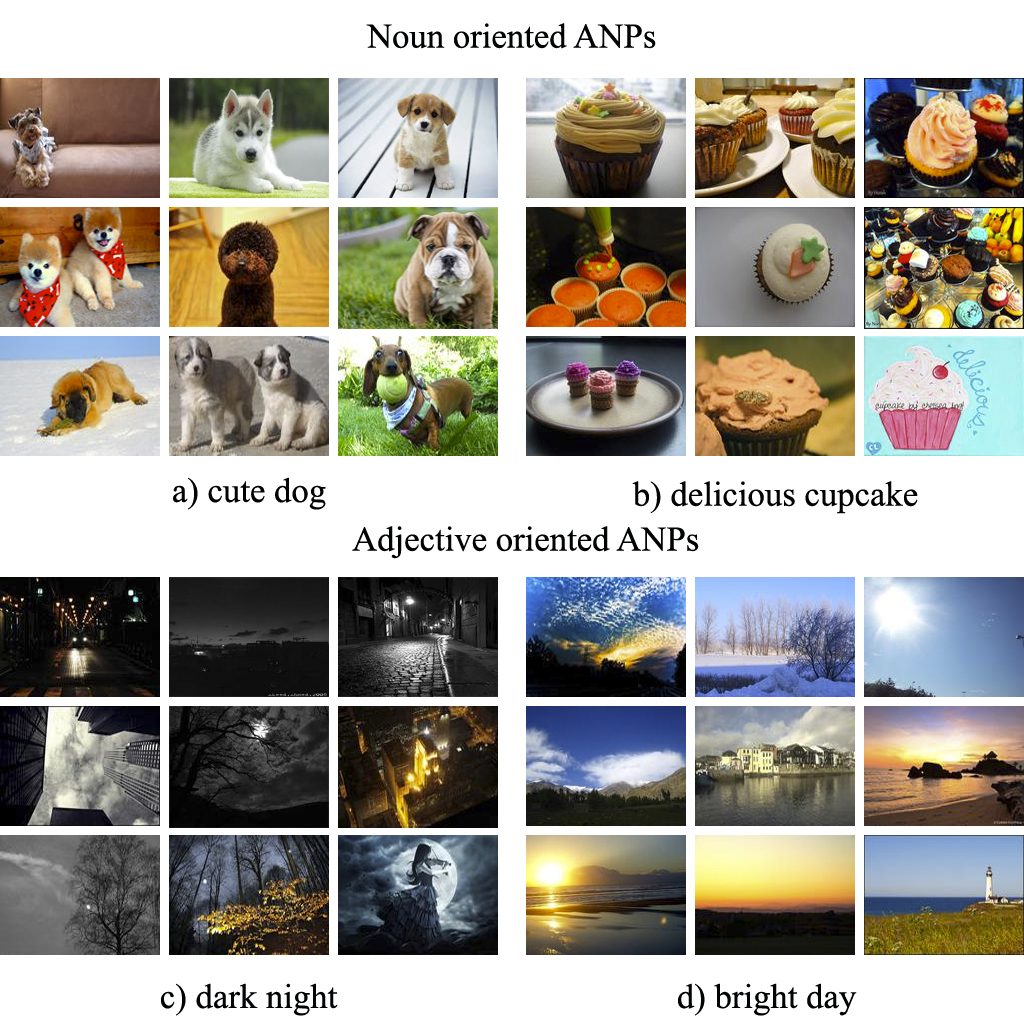}
\caption{Example of object oriented and scene oriented ANPs. The hypothesis of different contribution of the adjective or the noun depending on the ANP is subtended on the subjective different visual relevance of one or the other varying on the ANP. We distinguish between noun oriented ANPs (top row) and adjective oriented ANPs (bottom row).
}
\label{figure:hypothesis}
\end{figure}

Computers are acquiring increasing ability for understanding high level visual concepts such as objects and actions in images and videos, but often lack an affective comprehension of such content.
Technologies have largely obviated emotion from data, while neurology demonstrates how emotions are fundamental to human experience by influencing cognition, perception and everyday tasks such as learning, communication and decision-making \cite{lane2002cognitive}.

During the last decade, with the growing availability and popularity of opinion-rich resources such as social networks, the interest on the computational analysis of sentiment has increased. Every day, Internet users post and share billions of multimedia content in online platforms to express sentiment and opinions about several topics \cite{jou2016large}, which has motivated research on automated affect understanding for large-scale multimedia \cite{VSO, MVSO, jiang2014predicting}. The ability of analyzing and understanding this kind of information opens the door to behavioral sciences and applications such as brand monitoring or advertisement effect \cite{picard1997affective}. 

One of the main challenges for automated affect understanding in visual media is overcoming the \textit{affective gap} between low-level visual features and high-level affective semantics. Such task goes beyond overcoming the \textit{semantic gap}, i.e.~recognizing the objects in an image, and poses a challenging task in computer vision. In \cite{VSO}, adjective-noun pair (ANP) semantics are proposed as a mid-level representation that convey strong affective content while being visually detectable by traditional computer vision methods, e.g.~``happy dog", ``misty morning" or ``little girl". 


It has been argued that the noun in an ANP grounds the visual appearance of the concept, while the adjective works as a bias carrying most of the conveyed affect \cite{MVSO}. However, we hypothesize that for some ANPs the adjective may carry most of the visual cues that are key for its detection, as depicted by the examples in Figure \ref{figure:hypothesis}.
While the most salient visual cues for ``cute dog'' or ``delicious cupcake'' are related to ``dog'' and ``cupcake'', we expect that in other cases such as ``dark night'' or ``bright day'' the visual features related to ``dark'' and ``bright'' to contribute more in the detection of the ANP.
The analysis developed in this paper allows classifying between noun or adjective oriented ANPs by comparing the contribution of the adjective and noun concept in the final prediction.

The examples depicted in Figure\ref{figure:hypothesis} point at a second factor influencing the relative contributions between adjective and nouns.
If focused on the nouns only, it can be observed that some of them are related to the objects depicted in the images (eg. dog and cupcake), while other nouns are more related to the whole scene represented by the image (eg. night and day).
Our analysis will also discuss the behavior of adjective and noun contributions from this perspective.

The prediction of these adjective and noun structured labels has been traditionally addressed with single branch classifiers, ignoring the particular structure of these pairs. In order to verify our hypothesis, we propose fusing specialized adjective and noun detectors and then analyze their contribution by means of state of the art methods. The proposed two-stage training process allows to decompose the decision of the final classifier in terms of the contribution of different understandable concepts, i.e.~the adjectives and nouns in the dataset. Given these contribution results, a thorough analysis is performed in order to understand how the classifier leverages the information coming from each of the branches and shed some light into the particularities of the ANP detection task.

The contributions of this work include (1) a new ANP classifier architecture that provides comparable performance to the state of the art results while allowing interpretability, (2) a method to evaluate adjective and noun contributions on ANP prediction, and (3) an extended analysis on the contributions from a numerical and semantical point of view.




This paper is structured as follows.
Section \ref{sec:RelatedWork} reviews the related models in ANP prediction and the previous works on decomposing this task into adjective and noun classification.
We present our interpretable model \textit{ANPnet} in Section \ref{sec:architecture}, which is trained with the dataset and parameters described in Section \ref{sec:ExperimentalSetup}.
Accuracy results are presented in Section \ref{sec:prediction}, while the contributions in terms of adjective and nouns are discussed in \ref{sec:contributions}.
Final conclusions and discussions are contained in Section \ref{sec:Conclusions}.
Source code and models are available at \url{https://imatge-upc.github.io/affective-2017-musa2/}.

\section{Related Work}
\label{sec:RelatedWork}
Automated sentiment and emotion detection from visual media has received increasing interest by the multimedia community during the past years. These tasks have been addressed with traditional low-level feature based techniques, such as color features \cite{jia2012can}, SIFT-based Bag of Words \cite{li2012scaring} and aesthetic effects \cite{wang2012understanding}. Due to the success of deep learning techniques in reference vision benchmarks \cite{AlexNet, GoogLeNet,residual_learning}, Convolutional Neural Networks (CNN) and Recurrent Neural Networks (RNNs) have replaced and surpassed handcrafted features for affect-related tasks \cite{you2015robust,campos2017pixels,you2016robust}. 

Methods for detecting adjective-noun pairs (ANP) have strongly relied on state-of-the-art vision algorithms. Similarly to other computer vision tasks, early approaches based on handcrafted features \cite{Sentibank} were soon replaced with CNNs \cite{deepVSO,MVSO,MVSO_deeper}.
CNNs have proven their efficiency for large-scale image datasets \cite{AlexNet, GoogLeNet,residual_learning,huang2017densely}. DeepSentiBank \cite{deepVSO} presented the first application of CNNs for ANP prediction. 
MVSO detector-banks \cite{MVSO_deeper} showed performance improvement by using a more modern architecture, GoogLeNet \cite{GoogLeNet}, which also reduced the amount of parameters of the model. 
 
The multi-task nature of the ANP detection task has been exploited by using a fan-out architecture, where a first set of layers is shared for all tasks, and then splits in different network heads that specialize on each task \cite{jou2016deep}. Inter-task influence is increased through the use of cross-residual connections between the different heads. Despite this approach improves on the single-task models, the hierarchical structure of ANPs is not explicitly encoded in the architecture and the influence of the adjective and noun branches on the ANP detection lacks interpretability as compared to the approaches presented in this work.

Factorized nets \cite{narihira2015mapping} explicitly leverage the hierarchical nature of ANPs in the model architecture. Decomposing the ANP detection task into factorized adjective and noun classification problems allows the model to classify unseen adjective and noun combinations that are not available in the training set. However, the use of an $M$-dimensional latent space for adjectives and nouns complicates the interpretability of their combination in terms of understandable semantic concepts.

The task of sentiment analysis is addressed with Deep Coupled Adjective and Noun neural networks (DCAN) \cite{wang2016beyond} by learning a mid-level visual representation from two convolutional networks jointly trained with VSO. 
One of this networks is specialized in adjectives and the other one in nouns.
The learned representations shows superior performance in the task of sentiment analysis, but does not provide an interpretation about which concepts triggered the predictions.


The model proposed in this work follows a fan-in architecture which allows to understand and decompose the final classification in terms of the contribution of specific adjectives and nouns.


\section{ANPnet}
\label{sec:architecture}

\begin{figure}
\includegraphics[width=\linewidth]{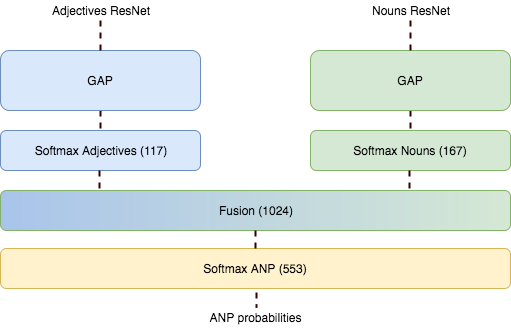}
\caption{Interpretable model for ANP detection. }
\label{fig:anp_arquitectures}
\end{figure}

This section presents \textit{ANPnet}, an interpretable architecture for ANP detection constructed by fusing the outputs of two specialized networks for adjective and noun classification. Given an input image, $x$, we estimate its corresponding Adjective, Noun and ANP labels, $y_{adj}$, $y_{noun}$ and $y_{ANP}$, as
\begin{align}
	&\hat{y}_{adj} = f_{adj} \left(x \right) \\
    &\hat{y}_{noun} = f_{noun} \left(x \right) \\
    &\hat{y}_{ANP} = g \left( \hat{y}_{adj}, \hat{y}_{noun} \right) \label{eq:fusion}
\end{align}   
where $f_{adj}$, $f_{noun}$ and $g$ are parametrized by neural networks, namely AdjNet, NounNet and the Fusion network. We aim at studying the contribution of the different adjectives and nouns by analyzing the behavior of $g$ with respect to its inputs. The method for computing such contributions is described in Section \ref{sec:contributions}.


The architecture for the specialized networks are based on the well-known ResNet-50 model \cite{residual_learning}. Residual Networks are convolutional neural network architectures that introduce residual functions with reference to the layer inputs, achieving better results than their non-residual counterparts. All residual layers use ``B option'' shortcut connections as described in \cite{residual_learning}, where projections are only used when matching dimensions ($1 \times 1$ convolutions with stride 2) and other shortcuts are identity.
This architecture represents a good trade-off between classification accuracy and computational cost. Besides, it allows comparison in terms of accuracy with the ResNet-50 network trained for ANP detection in \cite{jou2016deep}.


The last layer in ResNet-50, originally predicting the 1,000 classes in ImageNet, is replaced in AdjNet and NounNet to predict, respectively, the adjectives and nouns considered in our dataset.
Each of these two networks is trained separately with the same images and the corresponding adjective or noun labels. The probability ouputs of AdjNet and NounNet are fused by means of a fully connected neural network with a ReLU non-linearity. On top of that, a softmax linear classifier predicts the detection probabilities for each ANP class considered in the model.


Figure \ref{fig:anp_arquitectures} depicts the deeper layers of ANPnet, showing how AdjNet and NounNet are fused by a fully connected layer of 1,024 neurons.
This size is chosen to allow the network to learn sparse representations between the neighboring layers of smaller sizes. Inputs to the fusion layer are whitened by computing the mean and standard deviation for all the samples in the training set. The number of output neurons in AdjNet and NounNet is determined by the number of adjective and noun classes in our dataset, which were 117 and 167 respectively.

Next sections present how ANPNet was trained to simultaneously provide competitive and interpretable results.





\section{Experimental setup}
\label{sec:ExperimentalSetup}

The ANPnet network presented in Section \ref{sec:architecture} was trained with a subset of the Visual Sentiment Ontology (VSO) \cite{VSO} dataset.
Firstly, AdjNet and NounNet were trained independently for adjective and noun predictions, respectively, and in a second stage the fusion layers were trained to predict the ANPs in the input image.
This section describes in detail the dataset used and the training parameters of the whole architecture.

\subsection{Dataset}
The presented work uses a subset of the Visual Sentiment Ontology (VSO) \cite{VSO} dataset, the same part used in \cite{jou2016deep} to facilitate the comparison in terms of accuracy.

The original VSO dataset contains over 1200 different ANPs and about 500k images retrieved from the social multimedia platform \textit{Flickr} \footnote{https://www.flickr.com}.  
Those images were found by querying \textit{Flickr} search engine with keywords from the \textit{Plutchik's Wheel of Emotions} \cite{d1957tannenbaum}, a well-known emotion model derived from psychological studies.
This wheel contains 24 basic emotions, such as \textit{joy}, \textit{fear} or \textit{anger}. 
The discovery of affective-related ANPs was based on their co-occurrences with the emotion tags. 
The initial list of ANP candidates was manually filtered in order to ensure semantics correctness, sentiment strength and popular usage on Flickr. 
Finally, each resulting ANP concepts was used to query again \textit{Flickr} and build this way a collection of images and metadata associated to the ANP. 

The full VSO dataset presents certain limitations already pointed out in \cite{jou2016deep}.
First, some adjective-noun pair concepts are singletons and do not share any adjectives or nouns with other concept pairs. Also, some nouns are massively over-represented and there are far less adjectives to compose the adjective-noun pairs. 
Due to these drawbacks, our experiments are based on a subset of VSO build according to the more restrictive constraints proposed in \cite{jou2016deep}. 
The considered subset of ANPs satisfy the following conditions: (1) the adjective is paired with at least two more different nouns, (2) nouns that are not overwhelmingly biasing and abstract, and (3) all ANPs are related with 500 or more images.
The final VSO subset contains 167 nouns and 117 adjectives that form 553 adjective-noun pairs over 384,258 Flickr images. A stratified 80-20 split is performed, resulting in 307,185 images for training and 77,073 for test. The partition used in our experiments is the same for which results in \cite{jou2016deep} are reported\footnote{Dataset splits were obtained through personal communication with the authors of \cite{jou2016deep}.}.


\subsection{Training}
\label{ssec:training}

All CNNs were trained using stochastic gradient descent with momentum of 0.9, on batches of 128 samples and a learning rate of 0.01. Data augmentation, consisting in random crops and/or horizontal flips on the input images, together with $\ell_2$-regularization with a weight decay rate of $10^{-4}$ is used in order to reduce overfitting. When possible, layers were initialized using weights from a model pre-trained on ImageNet \cite{deng2009imagenet}. Otherwise, weights were initialized following the method proposed by Glorot and Bengio \cite{glorot2010understanding} and the initial value for the biases was set to 0.

The training is performed in two stages. First, AdjNet and NounNet are trained independently for the tasks of adjective and noun classification, respectively. The learned weights are then frozen in order to train the fusion network on top of the specialized CNNs. Thanks to this two-step training strategy, the inputs to the fusion layer become an intermediate and semantically interpretable representation. 

All experiments were run using two NVIDIA GeForce GTX Titan X GPUs and implemented with TensorFlow \cite{abadi2015tensorflow}.



\section{ANP Prediction}
\label{sec:prediction}

\begin{figure}
\includegraphics[width=\linewidth]{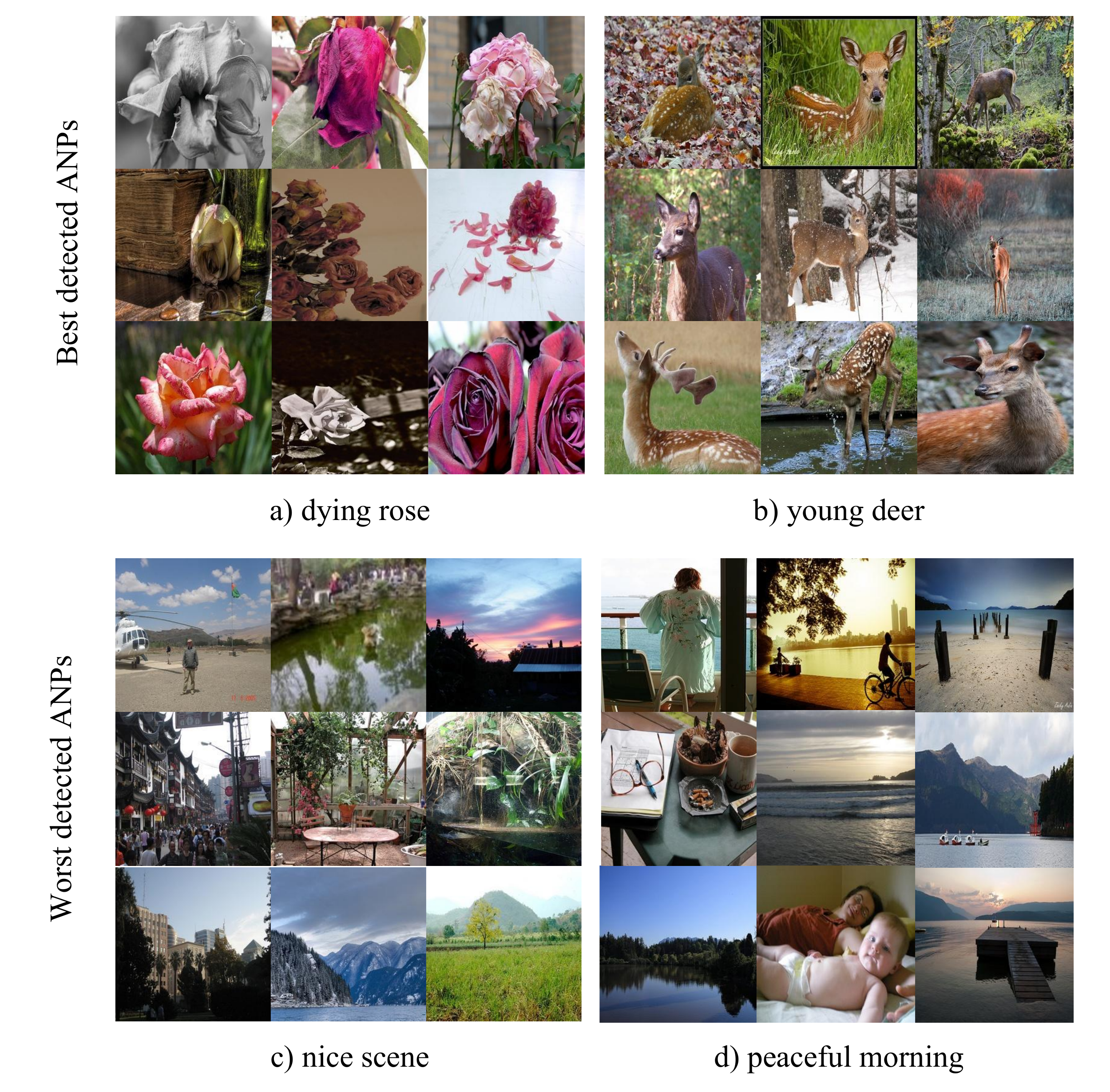}
\caption{Example of two of the best (a, b) and worst detected (c, d) ANPs. The visual variance on the ANP can be noticed with contrasting examples.}
\label{figure:best_worst_anps}
\end{figure}

The performance of our interpretable model in terms of ANP detection is presented in Table \ref{table:adj_noun_accuracy}.
This table shows a decrease of performance in terms of accuracy of ANPnet with respect to a ResNet-50 fine-tuned end-to-end for ANP prediction.
The table includes the results of two ResNet-50, the ones published in \cite{jou2016deep} and new ones obtained with the training parameters described in Section \ref{ssec:training}.
The similar values obtained by our model with respect to the ones reported in \cite{jou2016deep} confirm that the training hyperparameters were appropriate.
According to the baseline set by our ResNet-50 for ANP prediction, the loss of accuracy associated building ANPNet is of 3.8\% for top-1 accuracy and 4.7\% for top-5.

The loss is also present when comparing ANPnet with a \textit{non-interpretable} version of the same architecture for which the output layers of AdjNet and NounNet are also initialized randomly and trained. In this case the drop in accuracy is of 2.2\% when considering top-1 and 2.5\% for top-5.
With this setup, the network is not forced to use adjective and noun probabilities as an intermediate representation and has additional degrees of freedom to optimize for the target task of ANP classification. 
The decrease in accuracy of the interpretable model with respect to this latter configuration is then expected.
These results quantize the price to pay in terms of accuracy for making our model interpretable. 

\begin{figure}
\includegraphics[width=\linewidth]{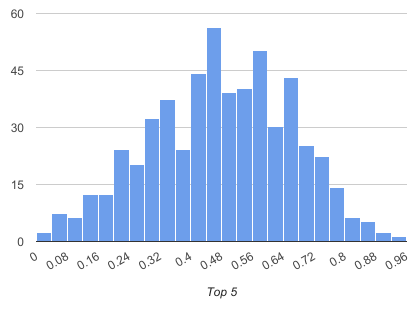}
\caption{Histogram of top-5 accuracy for ANP prediction with ANPnet. }
\label{fig:top5-histogram}
\end{figure}

A more general overview of the ANPnet performance in terms of top-5 accuracy is presented in Figure \ref{fig:top5-histogram} as the histogram of values considering the full test dataset.
The distribution of accuracies across the different ANPs follows a Gaussian-like distribution centered around the average score of 43.28\%.

\begin{table}[t]
\centering
\begin{tabular}{@{}lcccc@{}}
\toprule
\textbf{Model} 									& \textbf{Task}  & \textbf{Classes}	& \textbf{top-1} 	& \textbf{top-5} \\ 
\midrule
AdjNet \cite{jou2016deep}				& Adj			 	& 117				&	28.45      		& 57.87        	\\
AdjNet									& Adj			 	& 117				&	27.70      		& 57.00       	 \\
\midrule	
NounNet \cite{jou2016deep}				& Noun			 	& 167				&	41.64      		& 69.81        	\\
NounNet 								& Noun			 	& 167				&	41.50      		& 69.20        	\\
\midrule
ResNet-50 \cite{jou2016deep}					& ANP			 	& 553				&	22.68      		& 47.82        	\\
ResNet-50				  	 					& ANP			 	& 553				&	23.40      		& 48.20   		\\
Non-Interpretable 								& ANP			 	& 553				&	21.80      		& 46.00  \\
\textbf{ANPNet} 								& ANP			 	& 553				&	20.67      		& 43.28   		\\                                   
\bottomrule
\end{tabular}
\caption{Detection accuracy for adjective, nouns and ANPs, in \%.}
\label{table:adj_noun_accuracy}
\end{table}

The results in Table \ref{table:adj_noun_accuracy} show both accuracy in terms of top-1 and top-5 because ANP prediction can be highly affected by synonyms.
Adjective concepts such as ``smiling'' and ``happy'', or noun concepts like ``cat'' and ``animal'' are considered absolutely different in the accuracy metric, so relaxing its detection by considering the top-5 predictions may provide a metric that expresses better the obtained results \cite{MVSO_deeper}.
This observation motivates the use of top-5 accuracy as the reference metric for evaluating the correctness of the predictions in the remaining sections of this work.

The accuracy results in Table \ref{table:adj_noun_accuracy} also show how adjectives are more difficult to detect than nouns, as accuracy values are lower, even distinguishing among a larger number of classes.
There are two reasons for this gap of performance.
The first one is that adjectives usually describe more abstract concepts than nouns, with a larger associated visual variance. 
For example, there may be a wide range of visual features required to describe the concept ``happy''. 
The second reason is that ResNet-50 was initially trained for object classification on ImageNet, a type of concepts that are associated to nouns.

A closer look at the results allows to distinguish which of the 553 considered ANPs can be better detected and which ones present more problems.
Table \ref{table:BestVsWorsANPs} presents the ANPs with best and worst top-5 accuracy predictions with ANPNet, comparing their accuracy per ANP with the individual accuracies of their composing adjective and nouns.
Qualitative results of two of the best and worst detected ANPs are depicted in Figure \ref{figure:best_worst_anps}.

These results show how the best detected ANPs correspond to \textit{object-oriented} nouns, i.e.~well defined entities with a low variation from a visual perspective and usually represented by a localized region withing the image. These would the the cases of ``river'', ``deer'' or ``mushrooms''. 
On the other hand, the worst predictions are associated to \textit{scene-oriented} ANPs, which depict more abstract concepts and thus have a larger visual variance, such as ``places'', ``view'' or ``scene'' itself.
The top-5 accuracy tends to be significantly better in the case of object-oriented nouns than in the case of scene-oriented ones.
This difference in performance may be related to using a ResNet-50 pre-trained on the ImageNet dataset for AdjNet and NounNet. ImageNet is a dataset build for object classification, so our model is more specialized in these type of classes than those related to scenes \cite{zhou2014learning}.

\begin{table*}[htp]
\centering
\begin{tabular}{@{}|l|ccc|l|ccc|@{}}
\toprule
\multicolumn{1}{|l|}{} & \multicolumn{3}{|c|}{\textbf{Top-5 accuracies for best ANPs}}    &  \multicolumn{1}{c|}{} & \multicolumn{3}{c|}{\textbf{Top-5 accuracies for worst ANPs}}     \\ 
\multicolumn{1}{|l|}{}       					& Adj & Noun & \textbf{ANP}  &  \multicolumn{1}{c|}{} & Adj & Noun & \textbf{ANP}     \\ 
\midrule
\multicolumn{1}{|l|}{\textbf{gentle river}}      & 72.22 & 55.97 & 92.45     & \multicolumn{1}{l|}{\textbf{abandoned places}}     	& 84.58 & 33.93 & 8.21         	\\
\multicolumn{1}{|l|}{\textbf{tiny bathroom}}   	& 70.82 & 88.35 & 91.26     & \multicolumn{1}{l|}{\textbf{beautiful landscape}}     & 60.49 & 42.68 & 7.41        		\\
\multicolumn{1}{|l|}{\textbf{young deer}}    	& 52.49 & 89.04 & 89.81     & \multicolumn{1}{l|}{\textbf{beautiful earth}}    		& 60.49 & 5.00 & 5.71				\\
\multicolumn{1}{|l|}{\textbf{wild deer}}    	 	& 64.34 & 89.04 & 86.49     & \multicolumn{	1}{l|}{\textbf{charming places}}       	& 4.46 & 33.93 & 5.36				\\
\multicolumn{1}{|l|}{\textbf{misty road}} 		& 79.42 & 80.16 & 86.49     & \multicolumn{1}{l|}{\textbf{bad view}}       			& 38.83 & 67.86 & 5.22             \\	
\multicolumn{1}{|l|}{\textbf{dying rose}}    	& 68.12 & 80.07 & 85.00     & \multicolumn{1}{l|}{\textbf{peaceful morning}}       	& 26.22 & 53.18 & 4.96				\\
\multicolumn{1}{|l|}{\textbf{icy grass}}       	& 78.29 & 69.31 & 84.16     & \multicolumn{1}{l|}{\textbf{peaceful places}}      	& 26.22 & 33.93 & 4.59             \\
\multicolumn{1}{|l|}{\textbf{tiny mushrooms}}    & 70.82 & 82.52 & 84.00     & \multicolumn{1}{l|}{\textbf{nice scene}}     			& 45.20 & 25.65 & 4.39             \\
\multicolumn{1}{|l|}{\textbf{golden statue}}    	& 64.56 & 77.12 & 83.61     & \multicolumn{1}{l|}{\textbf{serene scene}}       		& 18.93 & 25.65 & 3.60             \\
\multicolumn{1}{|l|}{\textbf{empty train}}       & 69.72 & 65.42 & 76.60     	& \multicolumn{1}{l|}{\textbf{bright sky}}       		& 52.84 & 67.93 & 3.20             \\
\bottomrule
\end{tabular}
\caption{Top-5 accuracies for the best and worst detected ANPs, together with the top-5 accuracies of their composing adjectives and nouns.} 
\label{table:BestVsWorsANPs}
\end{table*}

ANPNet allows a finer analysis under the form of a \textit{co-detection matrix}.
The contents of the matrix provide the percentage of images for which the adjective, noun or ANP are correctly detected (columns) among those images for which the adjective, noun or ANP has been correctly predicted (rows).
As previously stated, a detection is considered correct when the ground truth label is among the top-5, whether adjective, noun or ANP.
The diagonal of the co-detection matrix corresponds to 100\%, as it represents the ratio of the whole set of detected adjective, noun or ANP with respect to itself.
The rest of values do not necessarily need to meet the 100\% as, for example, a correct detection of an ANP among the top-5 predictions does not also imply that the composing adjective or noun was among the top-5 predicted adjectives or nouns.
The co-detection matrix allows to study the correlation between ANPs and their composing parts, providing insights about how a correct detection of adjectives and nouns is related to a correct detection of an ANP.

The co-detection matrix of ANPNet is presented in Table \ref{tab:codetection-matrix}.
Its results indicate that the detection of an ANP also implies in many cases a correct detection of the adjective, in 87.47\% of the cases, and of the noun, in 95.76\% of the cases. On the other hand, a correct detection of the adjective or noun does not necessarily imply a correct detection of the ANP, as the ANP top-5 accuracy for these cases is 66.38\% for adjectives and 59.87\% for nouns.

In addition, the matrix also indicates that a detection of the adjective is also related to a correct prediction of the noun in 84.58\% of the samples, but the inverse situation only corresponds to 69.67\% of the considered images. 



\begin{table}[ht]
\centering
\begin{tabular}{cccc}
\multicolumn{1}{c|}{}			& Adj 		& Noun 		& ANP		\\ \hline
\multicolumn{1}{c|}{Adj} 		& 100.00 	& 84.58 	& 66.38	\\ 
\multicolumn{1}{c|}{Noun} 		& 69.67		& 100.00 	& 59.87	\\ 
\multicolumn{1}{c|}{ANP} 		& 87.47		& 95.76		& 100.00 	\\ 
\end{tabular}
\caption{Adjective, Noun and ANP co-detection matrix. The contents of the matrix provide the percentage of images for which the adjective, noun or ANP are correctly detected (columns) among those images for which the adjective, noun or ANP has been correctly predicted (rows).}
\label{tab:codetection-matrix}
\end{table}

\section{Adjective and Noun Contributions}
\label{sec:contributions}

Understanding neural networks has attracted much research interest. For CNNs in particular, most methods rely on extracting the most salient points in the original image space that triggered a particular decision. The particularities of the ANPNet model presented in Section \ref{sec:architecture} allows to interpret the ANP predictions obtained in Section \ref{sec:prediction} in terms of adjective and noun contributions.

We adopted Deep Taylor Decomposition \cite{montavon2017explaining} to compute the contribution of each element in $\hat{y}_{adj}$ and $\hat{y}_{noun}$ to the final ANP prediction, $\hat{y}_{ANP}$ (Equation \ref{eq:fusion}). This method consists in computing and propagating relevance backwards in the network (i.e.~from the output to the input), where layer-wise relevance is approximated by a domain-specific Taylor decomposition. Two different rules are then derived for ReLU layers, namely the $z^+-$rule and $z^B-$rule, which apply to layers with unbounded and bounded inputs, respectively \cite{montavon2017explaining}. In our model, $z^B-$rule is used for the relevance model between the fully connected layer in the Fusion network and the bounded Adjective and Noun probabilities, whereas $z^+-$rule is used otherwise.

\subsection{Adjective Noun Ratio}

\begin{table*}[htp]
\centering
\begin{tabular}{@{}|l|ccccc|ccc|@{}}
\toprule
				& \multicolumn{5}{c|}{\textbf{Adjective-to-Noun Ratio (ANR)}}    			& \multicolumn{3}{c|}{\textbf{Top-5 Accuracy}}     \\ 
			   	& \multicolumn{4}{c|}{Correct predictions} 		& \multicolumn{1}{c|}{All top-5 predictions} & \multicolumn{3}{c|}{}  \\ 
 			 	& \textbf{ANP} 	& ANP + Adj	& ANP + Noun 	& ANP + Adj + Noun & \multicolumn{1}{|c|}{ANP}		& Adj 	& Noun 	& ANP \\ 
\midrule	
sexy model 		& 1.161  		& 1.162	 	& 1.163     	 & 1.163 		& \multicolumn{1}{|c|}{1.122} 				& 76.52  & 62.77   	&	59.63  \\
misty trees 	& 1.139  		& 1.140	 	& 1.138     	 & 1.139 		& \multicolumn{1}{|c|}{1.146} 				& 79.42  & 71.74   	&	71.88  \\
abandoned places& 1.121  		& 1.121	 	& 1.033     	 & 1.033 		& \multicolumn{1}{|c|}{1.018} 				& 84.58  & 33.93   	&	8.21  \\
sexy body 	 	& 1.118  		& 1.118	 	& 1.117     	 & 1.117 		& \multicolumn{1}{|c|}{1.110} 				& 76.52  & 57.89   	&	56.44  \\
wild horse 		& 1.117  		& 1.117	 	& 1.116     	 & 1.117 		& \multicolumn{1}{|c|}{1.109} 				& 54.04  & 88.50   	&	58.06  \\
\hline
innocent eyes 	& 0.787 	& 0.788	 		& 0.787     	 & 0.788 	& \multicolumn{1}{|c|}{0.788}				& 43.23  	& 76.44   	&	16.07  \\
incredible view & 0.785 	& 0.786	 		& 0.785     	 & 0.786 	& \multicolumn{1}{|c|}{0.809}				& 30.71  	& 67.86   	&	39.02  \\
tired eyes 		& 0.776 	& 0.778	 		& 0.776     	 & 0.788 	& \multicolumn{1}{|c|}{0.784}				& 56.13  	& 76.44   	&	37.50  \\
laughing baby 	& 0.769 	& 0.769	 		& 0.769     	 & 0.769 	& \multicolumn{1}{|c|}{0.773}				& 72.57  	& 83.74   	&	69.03  \\
chubby baby 	& 0.764 	& 0.764	 		& 0.764     	 & 0.764 	& \multicolumn{1}{|c|}{0.786}				& 48.00  	& 83.74   	&	45.60  \\
\bottomrule
\end{tabular}
\caption{Highest and lowest ANRs computed for all top-5 predictions and only for the correct predictions among the top-5.}
\label{table:anr}
\end{table*}


This first analysis explores the nature of different ANPs depending on how much their prediction is influenced by the adjective and noun classes that compose it.
For this purpose, we define the \textit{Adjective-to-Noun Ratio} (ANR) as the normalized contribution of the adjectives with respect to the nouns during the prediction of an ANP.
These normalized contributions are computed by summing the individual contributions of all adjectives and nouns considered by AdjNet and NounNet, and normalizing them by the total amount of adjective and nouns, respectively.
Based on this definition, a uniform distribution of  activations at the 117 outputs of AdjNet and another uniform distribution of activations at the 167 outputs of NounNet will result in an ANR equal to the unit.


We present two types of analysis from the ANR perspective: when considering a correct predictions of the ANP and its composing adjective and nouns, and when considering every prediction in the top-5 for every image. 
The ANRs of the ANP with highest and lowest ANPs are presented in Table \ref{table:anr}, together with their top-5 accuracy values for the prediction, which allows interpreting the ANRs values in the context of the predictability of the adjective, noun and ANP.

\subsubsection{Average ANR per correct predictions}
\label{ssec:ANRcorrect}

A first analysis considers high quality predictions from two perspectives: focusing on the correctness of the ANP only, or requiring a correct prediction the adjective and/or noun as well.
As in previous sections, a prediction is considered correct if the ground truth adjective, noun or ANP is among the top-5 prediction.
This analysis corresponds to the columns 2 to 5 in Table \ref{table:anr}.

A first observation is that ANR takes values higher or lower than one, indicating that some ANP predictions are more influenced by the adjective branch of ANPnet, while others are more influenced by the noun branch.
This diversity allows us to classify ANPs between \textit{adjective oriented} or \textit{noun oriented} ANPs, depending on whether the ANR is higher or lower than one, correspondingly.

A second observation indicates that the difficulty on predicting the adjective favors most of the ANPs to be noun oriented. We can observe in table \ref{table:anr} that all the five ANPs with lowest ANR (noun oriented) have an adjective accuracy prediction lower than the adjective predictions from the five ANPs with highest ANR. 

Finally, filtering the results by forcing a correct prediction of the adjective and/or noun shows almost no impact in the final estimated ANR.
This lack of variation can be explained by the co-detection matrix previously presented in Table \ref{tab:codetection-matrix}.
The third row of the matrix indicates that, when an ANP is correctly detected, the adjective is also well predicted in 87.47\% of the cases and the noun in 95.76\%.
This allows a small variation when samples are filtered to force that adjective and/or noun must also be detected.

\subsubsection{Average ANR per all images}
\label{ssec:ANRall}

The results of Section \ref{ssec:ANRcorrect} indicate how each ANP prediction is affected by the composing adjective and noun in the case of correct predictions.
We extend this analysis to a more generic approach were ground truth labels are not considered, so that all top-5 predicted ANP for each image are used in estimating the ANR values.
This analysis provides more samples because in this case each image in the dataset allows drawing five ANR values, one for each of the top-5 predicted ANPs.
In the case of Section \ref{ssec:ANRcorrect}, each image could only contribute in estimating the ANR of the ground truth ANP, and only in the case that the ANP were predicted among the top-5 ones.

The results of this analysis corresponds to the sixth column in Table \ref{table:anr}.
The estimated values show some slight variations with respect to the ANRs predicted over correct predictions only.
The conclusions about adjective and noun oriented ANP remain the same for the cases depicted in the table.


\subsection{Visually equivalent ANPs}

\begin{table}[htp]
\centering
\begin{tabular}{@{}cc|cc@{}}
\cmidrule(r){1-4}
\multicolumn{2}{c|}{\includegraphics[width=0.2\textwidth]{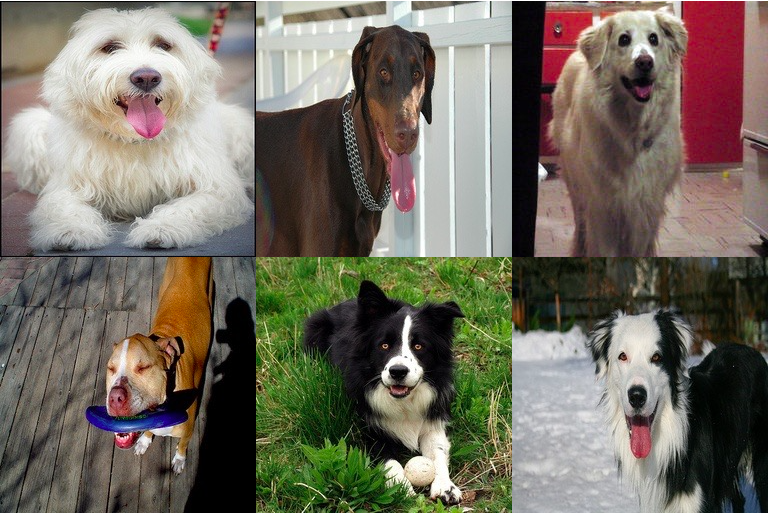} }
& \multicolumn{2}{c}{\includegraphics[width=0.2\textwidth]{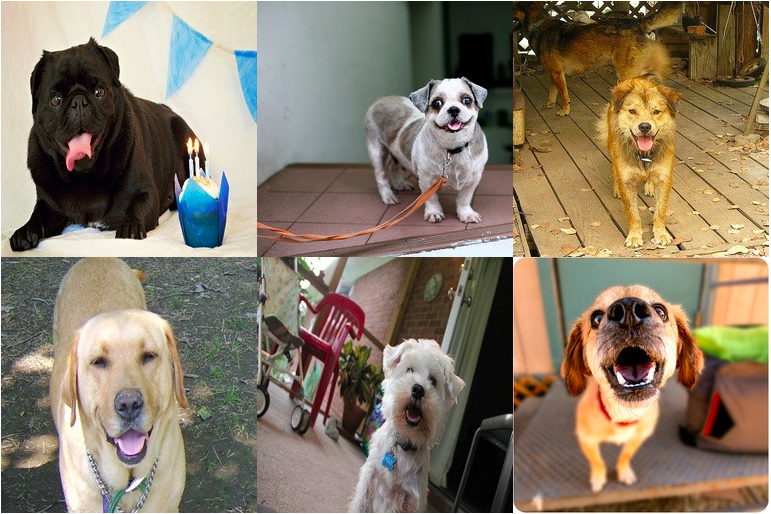}}        \\ \cmidrule(r){1-4} 
\multicolumn{2}{c|}{\textbf{Happy Dog}}           & \multicolumn{2}{c}{\textbf{Smiling Dog}}                  \\ \cmidrule(r){1-4}
\textbf{top-5 adjectives} & \textbf{top-5 nouns} & \textbf{top-5 adjectives} & \textbf{top-5 nouns}  \\ \cmidrule(r){1-4} 
happy         & dog        & smiling         & dog                 \\
smiling       & animals    & happy           & eyes                \\
friendly      & pets        & friendly        & pets               \\
playful       & grass       & funny           & blonde                 \\
funny         & eyes        & playful          & animals          \\ \\ \cmidrule(r){1-4}
\cmidrule(r){1-4}
\multicolumn{2}{c|}{\includegraphics[width=0.2\textwidth]{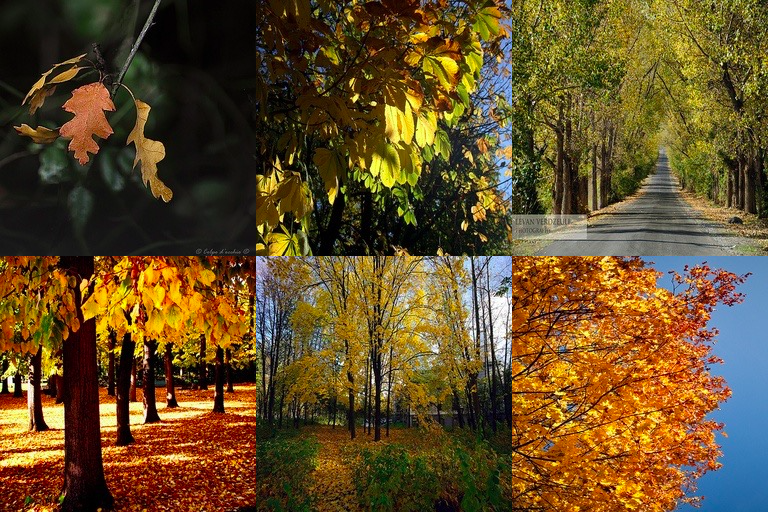} }
& \multicolumn{2}{c}{\includegraphics[width=0.2\textwidth]{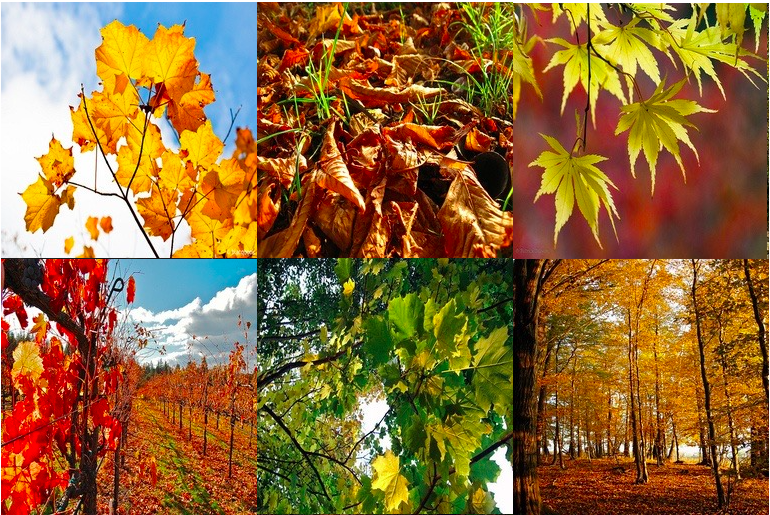}}        \\ \cmidrule(r){1-4} 
\multicolumn{2}{c|}{\textbf{Golden Autumn}}           & \multicolumn{2}{c}{\textbf{Golden Leaves}}                  \\ \cmidrule(r){1-4}
\textbf{top-5 adjectives} & \textbf{top-5 nouns} & \textbf{top-5 adjectives} & \textbf{top-5 nouns}  \\ \cmidrule(r){1-4} 
golden                   & autumn                  & golden            & leaves                 \\
sunny                 & leaves               	& sunny             & autumn                \\
colorful                   & trees             & falling           & trees               \\
falling            		 	& sunlight            	& colorful             & sunlight                 \\
bright                & tree            			& bright           & tree          \\ \cmidrule(r){1-4}
\end{tabular}
\caption{Pairs of ANPs with identical top-5 adjective and noun contributions.}
\label{table:semantic_contribution_comparison}
\end{table}

The interpretation of ANP predictions in terms of their contributing adjectives and nouns permits a novel description approach for ANPs.
We define as \textit{visually equivalent} those ANPs whose top-5 adjective and noun contributions are identical.
Table \ref{table:semantic_contribution_comparison} contains two pairs of visually equivalent ANPs. 
In the first example, ``happy dog'' and ``smiling dog'' have an identical noun and very similar adjective from a semantic perspective, while the second case presents the opposite situation, in which the same ``golden'' adjective is used to build the ``golden autumn'' and ``golden leaves'' ANPs.
In both examples their two sets of top-5 contributing adjectives and nouns is identical, but not in the exact order.
Also, from a visual perspective, the presented examples depict how the two ANPs are visually described by the same class of images.

A more extensive list of visually equivalent ANPs is provided in Table \ref{table:semantic_contribution_comparison}, together with the ANR of each ANP.
These visually equivalent ANPs also share in all cases the adjective or the noun.
Notice how they also present similar ANR values, an expected behavior as the most contributing adjectives and nouns are the same in each member of the pair.

These results show how our interpretable model is able to identify equivalent ANPs, which are a common case given the subjectivity and richness of an affective-aware dataset.
These observations reinforce the choice of a top-5 accuracy metric instead of a more rigid top-1, as the boundaries between classes are very dim and often overlap.

\begin{table}[htp]
\centering
\begin{tabular}{@{}lc|lc@{}}
\multicolumn{1}{c}{ANP}							& ANR	& \multicolumn{1}{c}{ANP}				& ANR \\
\toprule
ancient architecture		& 1.044	& ancient building	& 1.075 \\
dead fly					& 1.045	& dead bug			& 1.080 \\
traditional architecture    & 1.027	& traditional house & 1.004	\\
dry tree					& 0.962 & dying tree		& 0.832 \\
tiny boat					& 0.924 & little boat		& 0.909 \\
weird bug					& 0.925	& ugly bug			& 0.964	\\
heavy clouds				& 0.921	& dark clouds		& 0.948	\\
beautiful clouds			& 0.920 & beautiful sky		& 0.895	 \\
angry cat					& 0.820 & evil cat			& 0.816	\\
\end{tabular}
\caption{Pairs of Visually Equivalent ANPs}
\label{table:equivalent_anps}
\end{table}
\subsection{Related Adjectives and Nouns}

\begin{table}[htp]
\centering
\begin{tabular}{@{}cc|cc@{}}
\cmidrule(r){1-4}
\multicolumn{2}{c|}{\includegraphics[width=0.2\textwidth]{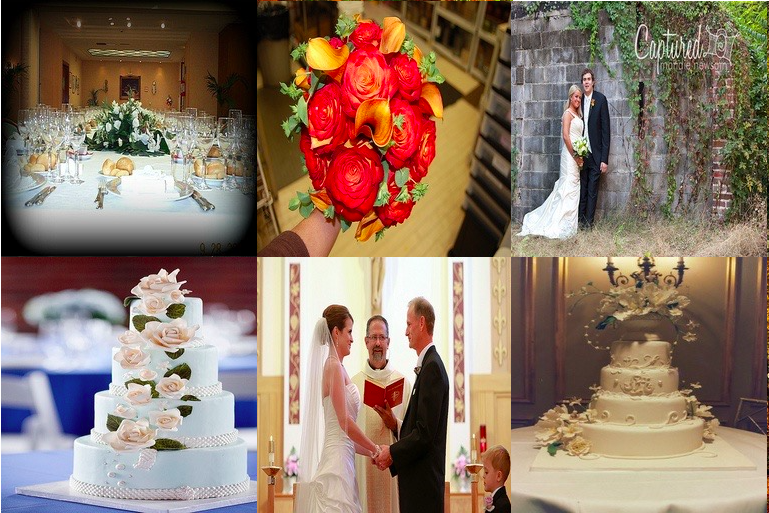} }
& \multicolumn{2}{c}{\includegraphics[width=0.2\textwidth]{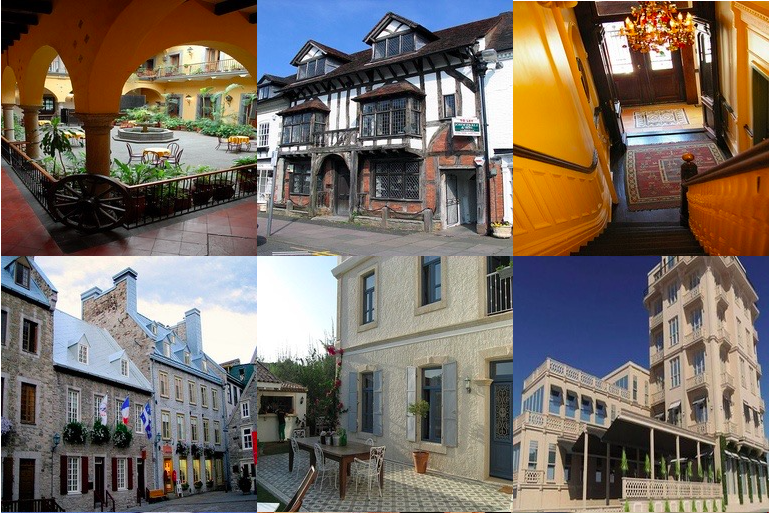}}        \\ \cmidrule(r){1-4} 
\multicolumn{2}{c|}{\textbf{a) Elegant Wedding}}           & \multicolumn{2}{c}{\textbf{b) Charming Places}}                  \\ \cmidrule(r){1-4}
\textbf{top-5 adjectives} & \textbf{top-5 nouns} & \textbf{top-5 adjectives} & \textbf{top-5 nouns}  \\ \cmidrule(r){1-4} 
elegant                      & wedding                  & charming                      & hotel                 \\
outdoor                 & cake               			 & comfortable                  & places                \\
fresh                   	 & rose             			  & excellent                    & house               \\
tasty            		 	& dress            			     & traditional             & home                 \\
delicious                & lady            					& expensive                    & food          \\ \\ \cmidrule(r){1-4}
\cmidrule(r){1-4}
\multicolumn{2}{c|}{\includegraphics[width=0.2\textwidth]{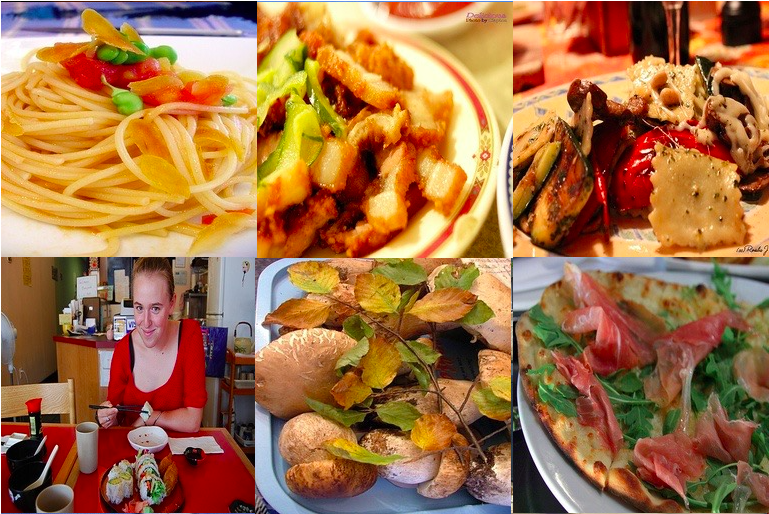} }
& \multicolumn{2}{c}{\includegraphics[width=0.2\textwidth]{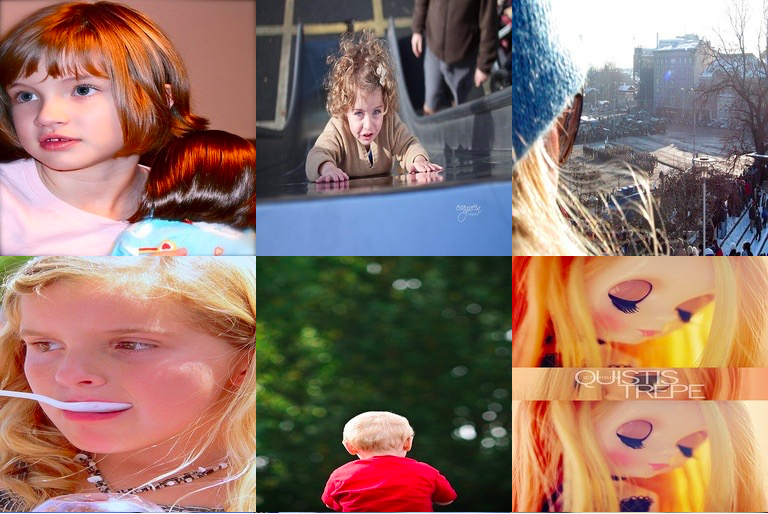}}        \\ \cmidrule(r){1-4} 
\multicolumn{2}{c|}{\textbf{c) Delicious Food}}           & \multicolumn{2}{c}{\textbf{d) Golden Hair}}                  \\ \cmidrule(r){1-4}
\textbf{top-5 adjectives} & \textbf{top-5 nouns} & \textbf{top-5 adjectives} & \textbf{top-5 nouns}  \\ \cmidrule(r){1-4} 
delicious                   & food                  & golden            & hair                 \\
traditional                 & cake               	& shiny             & lady                \\
excellent                   & mushrooms             & pretty           & blonde               \\
tasty            		 	& market            	& sexy             & sunlight                 \\
yummy                & drink            			& smooth           & girl          \\ \cmidrule(r){1-4}
\end{tabular}
\caption{Examples of co-occurring concepts on ANPs for the ANPs: a) "elegant wedding", b) "charming places", c) "delicious food" and d)  "golden hair". Notice how the most contributing concepts match the images on the dataset for a given ANP.}
\label{table:coocurrances}
\end{table}

The most contributing adjectives and nouns detected by ANPnet can also be used as semantic labels themselves.
This way, our model can detect in a single pass additional concepts related to the predicted ANP, with applications to image tagging, captioning or retrieval.


Table \ref{table:coocurrances} shows the top-5 related adjective and nouns of four ANPs. 
We verify that the top contributing adjective and nouns have correspondence with the image contents, by randomly picking six images in the dataset for the considered ANPs. 
In example a) it can be seen how ANPnet learned that the most related concepts for an ``elegant wedding'' scene are the names ``cake'', ``rose'', ``dress'' and ``lady'',  and the adjectives ``outdoor'', ``fresh'', ``tasty'' and ``delicious''. Notice the high contribution of food-adjectives as ``fresh'', ``tasty'' and ``delicious'', which apply to the wedding cake and wedding meal. 
In the example b) we show the highest contributions for a more scene-oriented ANP, as ``charming place''. We notice how the network has been able to learn that adjectives describing ``charming places'' are often also related to ``comfortable'', ``excellent'', ``traditional'' and ``expensive''; and that elements appearing on these types of scenes are ``hotel", ``house'', ``home'' and ``food''. Examples c) and d) show additional cases of adjectives and nouns that match the contents of some images. For example, to describe ``delicious food'' we could use adjectives as ``traditional'', ``excellent'', ``tasty'' or ``yummy''. And to describe ``golden hair'' images, other concepts that are related are: ``shiny'', ``pretty'', ``sexy'', ``smooth'', ``lady'', ``blonde'', ``sunlight'' and ``girl''. 

\section{Conclusions and Future Work}
\label{sec:Conclusions}

This work has presented ANPnet as an interpretable model capable of disentangling the adjective and noun contributions for the predictions of Adjective Noun Pairs (ANPs).
This tools has allowed us to validate our hypothesis that the contribution of adjectives and nouns varies depending on each ANP and have introduced Adjective-to-Noun Ratio (ANR) as a measure to quantize it.

ANPnet is based on the fusion of two specialized networks for adjective and noun detection.
Keeping the interpretation of the model when fusing the two specialized networks has also provoked a loss of accuracy of the model, establishing a trade-off between interpretability and performance.
It has been observed that better detection accuracies are often associated to object-oriented nouns, while worse ones are related to scene-oriented nouns.
As future work, one may explore pre-training AdjNet and NounNet not only with an object-oriented dataset as ImageNet, but also with a scene-oriented one such as Places \cite{zhou2014learning}.


The unbalanced contributions of adjective and nouns in ANP predictions also allows a classification between adjective- and noun-oriented ANPs.
Adjective-oriented ANPs tend to be harder to detect because adjectives themselves are also harder to detect than nouns. 
As in the case of scene-oriented nouns, adjective-oriented ANPs may be difficult to predict because AdjNet and NounNet were pre-trained with the objects in ImageNet, a type of nouns.
In addition, qualitative results also indicate how adjective concepts are much more visually diverse than noun ones.

Our work has also shown how different ANPs may have the same top adjective and noun contributions, allowing the detection of visually equivalent ANPs. As a final analysis, we have shown how ANPnet can also be used to generate adjective and noun labels to enrich the semantic description of the images. 


The presented work, while focus on affective-aware ANPs, could be extended to any other problem of adjective and noun detection, and even to more complex cases with more composing concepts.
Our interpretable model aims at contributing in the field of better understanding why deep neural networks produce their predictions in terms of intermediate activations with a straightforward semantic interpretation. In the other hand, relationship between concepts can be used in order to modify network loss when training.

As future work, the accuracy gap between the interpretable model and the baseline should be reduced. An option to do that is by introducing multi-task and weight sharing between AdjNet and NounNet to reduce the number of parameters.
A different architecture were the detection of an adjective or noun could be conditioned by the prior detection of the other (or viceversa) could reduce the visual variance in the detection and help improving its performance.
Finally, the introduction of external knowledge bases could be explored to explore prior knowledge on the domain.

The source code and models used in this work are publicly available at \url{https://imatge-upc.github.io/affective-2017-musa2/}.

\begin{acks}
D{\`e}lia Fern{\'a}ndez is funded by contract 2017-DI-011 of the Industrial Doctorate Programme of the Government of Catalonia.
This work was partially supported by the Spanish Ministry of Economy and Competitivity under contracts TIN2012-34557 by the BSC-CNS Severo Ochoa program (SEV-2011-00067), and contracts TEC2013-43935-R and TEC2016-75976-R. It has also been supported by grants 2014-SGR-1051 and 2014-SGR-1421 by the Government of Catalonia, and the European Regional Development Fund (ERDF). 
We gratefully acknowledge the support of NVIDIA Corporation for the donation of GPUs used in this work.
\end{acks}

\bibliographystyle{ACM-Reference-Format}
\bibliography{references} 

\end{document}